\documentclass[research]{fcs}

\usepackage{amssymb}
\usepackage{utfsym}
\usepackage{times}
\usepackage{latexsym}
\usepackage{lineno}
\usepackage{pifont}
\usepackage{float}
\usepackage{times}
\usepackage{latexsym}
\usepackage{times}
\usepackage{latexsym}
\usepackage{multirow}
\usepackage{url}
\usepackage{tabularx}
\usepackage{latexsym}
\usepackage{balance}
\usepackage{bm}
\usepackage{amssymb}
\usepackage{amsmath}
\usepackage{url}
\usepackage{makecell}
\usepackage{multirow}
\usepackage{graphicx} 
\usepackage{color}
\usepackage{colortbl}
\usepackage{textcomp}
\usepackage{CJK}
\usepackage{enumitem}
\usepackage{booktabs}
\usepackage{microtype}
\usepackage{arydshln}
\usepackage{amsfonts}
\usepackage{wasysym}
\usepackage{pifont}
\usepackage[normalem]{ulem}
\usepackage[T1]{fontenc}
\usepackage[utf8]{inputenc}
\usepackage{microtype}
\usepackage{graphicx}
\usepackage{booktabs}
\usepackage{multirow}
\usepackage{makecell}
\usepackage{xcolor, soul}
\usepackage{wrapfig}
\usepackage[marginal]{footmisc}
\renewcommand{\thefootnote}{}

\definecolor{light_blue}{rgb}{0.80,0.85,1.0}
\definecolor{light_red}{rgb}{1.0,0.85,0.80}
\definecolor{med_red}{rgb}{1.0,0.55,0.50}
\definecolor{light_orange}{rgb}{1.0,0.72,0.5}

\title{Achieving $>$97\% on GSM8K: Deeply Understanding the Problems Makes LLMs Better Solvers for Math Word Problems}
\shorttitle{Deeply Understanding the Problems Makes LLMs Better Solvers for Math Word Problems}
\author[1]{Qihuang~ZHONG$^*$}
\author[1]{Kang~WANG$^*$}
\author[1]{Ziyang~XU}
\author*[1]{Juhua~LIU}
\author[2]{Liang~DING}
\author*[1]{Bo~DU}
\address[1]{School of Computer Science, National Engineering Research Center for Multimedia Software, Institute of Artificial Intelligence, and Hubei Key Laboratory of Multimedia and Network Communication Engineering, \\Wuhan University, Wuhan 430072, China}
\address[2]{School of Computer Science, University of Sydney, Sydney 2006, Australia}
\fcssetup{
  received       = {month dd, yyyy},
  accepted       = {month dd, yyyy},
  corr-email     = {\{liujuhua, dubo\}@whu.edu.cn},
}

\begin{abstract}
 Chain-of-Thought (CoT) prompting has enhanced the performance of Large Language Models (LLMs) across various reasoning tasks. However, CoT still falls short in dealing with complex math word problems, as it usually suffers from three pitfalls: semantic misunderstanding errors, calculation errors, and step-missing errors. Prior studies involve addressing the calculation errors and step-missing errors, but neglect the semantic misunderstanding errors, which is the major factor limiting the reasoning performance of LLMs. To this end, we propose a simple-yet-effective method, namely \textit{Deeply Understanding the Problems} (DUP), to improve the LLMs' math problem-solving ability by addressing semantic misunderstanding errors. The core of our method is to encourage the LLMs to deeply understand the problems and extract the key problem-solving information used for better reasoning. Extensive experiments on 10 diverse reasoning benchmarks show that our DUP method consistently outperforms the other counterparts by a large margin. More encouragingly, DUP achieves a new SOTA result on the GSM8K benchmark, with an accuracy of 97.1\% under the zero-shot setting.\footnote{*~Equal contribution: Qihuang Zhong and Kang Wang contributed equally to this work.}
\end{abstract}

\begin{document}
\section{Introduction}
\label{sec:intro}
Despite the impressive performance of Large Language Models (LLMs) in diverse natural language understanding and generation tasks~\cite{brown2020language, touvron2023llama, openai2023gpt4,bai2023qwen}, they often suffer from sub-optimal reasoning abilities, which cannot be overcome solely by simply scaling up the model size~\cite{rae2021scaling, wang2022self_consistency}. To tackle this limitation, Wei~et~al.\cite{cot_wei} propose a few-shot Chain-of-Thought (CoT) prompting strategy, which prompts the LLMs to mimic the given step-by-step thought process a person might employ in solving a task. Such a simple strategy can significantly improve the reasoning ability of LLMs, and thus has attracted widespread attention in recent years.

\begin{figure}[ht]
	\centering
 	\includegraphics[width=0.45\textwidth]{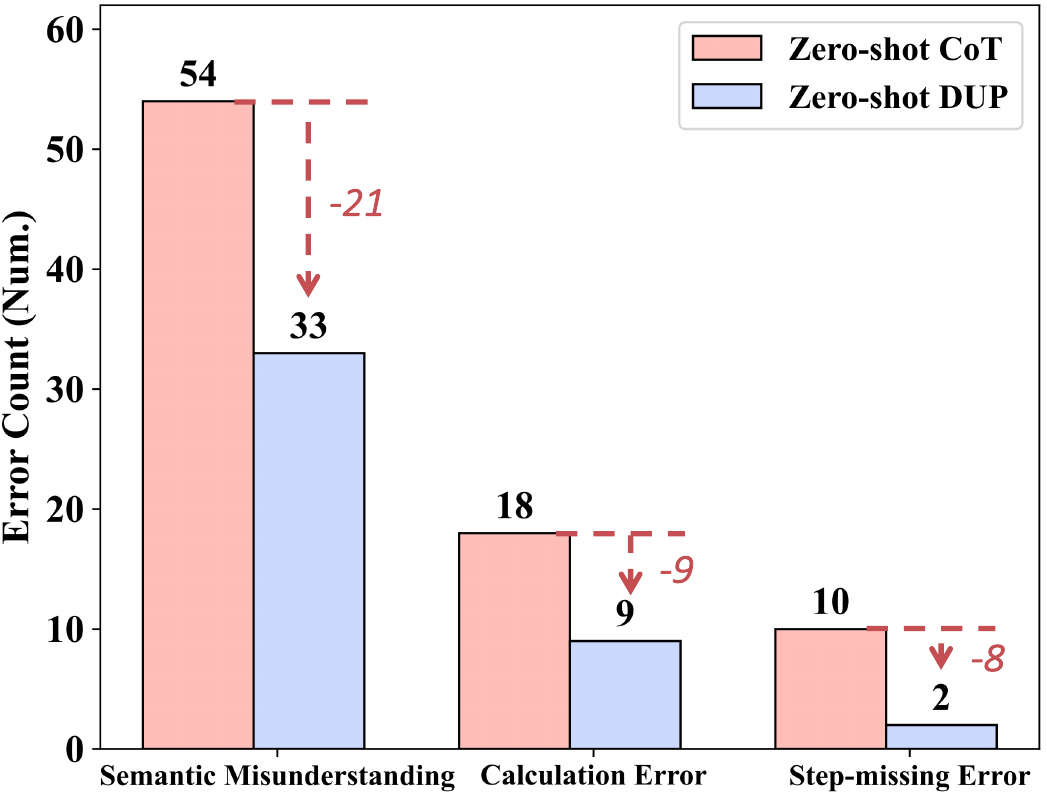} 
    \caption{\textbf{Error analysis of GSM8K problems with incorrect answers returned by zero-shot CoT and our DUP using GPT-3.5 LLM.} We randomly sample 300 GSM8K problems, and follow~\cite{cot_wei} and~\cite{wang2023planandsolve} to assign the ``Semantic Misunderstanding'', ``Calculation Error'' and ``Step-missing Error'' to each incorrect answer. The detailed prompts for error analysis are shown in Table~\ref{tab:prompt_error}. \textit{We see that our DUP method effectively reduces the errors among all types.}
 }
 \label{fig:intro_bar}
\end{figure}

Along this research line, many works focus on designing prompting strategies to enhance LLM's reasoning ability, such as Zero-shot CoT~\cite{kojima2022large}, Tree of Thought~\cite{gao2023pal}, Plan-and-Solve (PS) prompting~\cite{wang2023planandsolve}, and Complex CoT~\cite{fu2022complexity}. Although achieving remarkable progress, they still fall short in dealing with complex reasoning tasks, \textit{e.g.}, math word problems~\cite{gsm8k}. As stated by Wei~et~al.~\cite{cot_wei}, there are three main error types in the CoT-based reasoning: \textbf{\textit{semantic misunderstanding errors}}, \textbf{\textit{calculation errors}}, and \textbf{\textit{step-missing errors}}. In our preliminary experiments (as shown in Figure~\ref{fig:intro_bar}), we found that CoT has major errors in semantic understanding, which is the main factor limiting LLMs' reasoning performance. Prior studies~\cite{wang2023planandsolve,chen2023program} show that the carefully-designed prompting strategies can achieve much fewer calculation errors and step-missing errors, but still struggle to address the major semantic misunderstanding. Hence, there raises a question: \textit{\textbf{whether we can enhance the reasoning abilities of LLMs by reducing the semantic misunderstanding errors?}}

Intuitively, since complex math word problems usually contain content irrelevant to solving the task, LLMs might fail to identify the core question and extract the relevant problem-solving information, thus leading to semantic misunderstanding and poor performance. This can be also proved by the findings in psychology, as prior studies~\cite{hoyer1979effects,pasolunghi1999working} show that the irrelevant information may significantly decrease some children's and even adults' problem-solving accuracy. Hence, this inspires us that, \textit{it is crucial to enforce the LLMs to pay more attention to the core information and reduce the negative effects of irrelevant information.}

Motivated by this, we propose a simple-yet-effective method, namely \textit{Deeply Understanding the Problems} (DUP), to improve the LLMs' math problem-solving ability. The principle of our method is akin to the human learning process, \textit{i.e.}, for human students who receive a complex math word problem, they will read and comprehend the text of the problem, identify the core question that needs to be answered, and finally solve it with relevant problem-solving information. Specifically, DUP consists of three stages: \ding{182} Revealing the core question of the input problem; \ding{183} Extracting the problem-solving information relevant to solving the core question; \ding{184} Generating and extracting the final answer by combining the core question with problem-solving information. By doing so, LLMs can filter out irrelevant information and achieve better math reasoning performance.

We conduct a series of experiments on 10 reasoning datasets across math, commonsense, and symbolic reasoning benchmarks. The experimental results of GPT-3.5-Turbo~\cite{ouyang2022training} and GPT-4~\cite{openai2023gpt4} show that: 1) DUP consistently outperforms the other counterparts across all datasets by a large margin; 2) Zero-shot DUP can even outperform the few-shot methods on most reasoning datasets; 3) More encouragingly, DUP achieves new SOTA results on the popular GSM8K (\textbf{97.1\%}) and SVAMP (\textbf{94.2\%}). To summarize, our contributions are three-fold: (1) We reveal the underlying causes of semantic misunderstanding errors, and propose a simple yet effective approach (DUP) to effectively address the semantic misunderstanding and boost LLMs' math reasoning ability. (2) DUP is easy-to-implement and plug-and-play. It can be easily applied to various LLMs. (3) Extensive experiments show that DUP outperforms the other counterparts by a large margin, and achieves new SOTA results on GSM8K and SVAMP.

The rest of this paper is organized as follows. In Section~\ref{sec:related_work}, we briefly review the related works. In Section~\ref{sec:method}, we introduce our proposed DUP method in detail. Section~\ref{sec:experiments} reports and analyzes our experimental results. Lastly, we conclude our study in Section~\ref{sec:conclusion}.

\section{Related Works}
\label{sec:related_work}
\subsection{Reasoning with Large Language Models}
In recent years, we have witnessed numerous large language models (LLMs)~\cite{bert,brown2020language,palm,zhong2022toward,openai2023gpt4,touvron2023llama} that achieved tremendous success in the NLP community. Due to their emergent abilities~\cite{weiemergent}, these LLMs can achieve impressive few-shot and zero-shot performance in a variety of NLP tasks~\cite{hendrycksmeasuring,srivastava2023beyond,zhong2023can}. Specifically, as stated in~\cite{weiemergent}, with the scaling of model sizes, LLMs tend to show some emergent abilities, such as instruction-following~\cite{weifinetuned,sanhmultitask}, program execution~\cite{nyeshow}, and model calibration~\cite{kadavath2022language}. However, LLMs still struggle to provide stable and accurate answers when dealing with complex reasoning tasks~\cite{zhang2023cumulative}, such as math reasoning~\cite{gsm8k,svamp,aqua,addsub}, commonsense reasoning~\cite{commonsenseqa,strategyqa} and symbolic reasoning~\cite{cot_wei}. Recent works~\cite{yuan2023scaling,luo2023wizardmath,yu2023metamath,ho2023large,liu2023tinygsm} have shown that reasoning-augmented LLMs tuning with mathematical data can relatively improve reasoning ability. However, even with such progress, these models still perform poorly in complex reasoning problems. This indicates that there is still significant room for improving the LLMs' performance in complex reasoning tasks.

\subsection{Prompting Methods}
Despite the remarkable performance, the aforementioned training-based approaches usually require collecting large amounts of data and expensive computational costs, and may cause LLMs' universal ability to decrease. Hence, some works~\cite{cot_wei,kojima2022large} attempt to use cheaper prompting methods to strengthen the LLMs' reasoning abilities without additional training. Wei~et~al.\cite{cot_wei} are the first to propose the few-shot CoT prompting, which elicits a series of intermediate natural language reasoning steps before giving the final answer. So far, CoT prompting has been proven to significantly improve the reasoning capability of LLMs. Along this research line, numerous works ~\cite{zhou2023leasttomost,wang2023planandsolve,Yao2023TreeOT,zhang2022automatic,chen2022program,xu2023re} attempt to carefully design more effective prompting strategies to improve the reasoning ability of LLMs. Unfortunately, these prompt methods achieve remarkable performance, but still fail to deal with complex reasoning tasks, \textit{e.g.}, math word problems. As stated by Wei~et~al.~\cite{cot_wei}, the reasoning mistakes of LLMs can be classified into three categories: semantic misunderstanding errors, calculation errors, and step-missing errors. Some prior works~\cite{wang2023planandsolve,chen2023program} attempt to reduce these errors, and achieve some performance improvements. However, they mainly focus on the calculation errors and step-missing errors, but neglect the major semantic misunderstanding errors. That is, it is critical but under-explored to study how to address the semantic misunderstanding.

\textbf{Novelty of our work.} 
Our main contribution lies in discovering that CoT has major errors in semantic understanding, which is the main factor limiting LLMs' reasoning performance. To address this issue, we are inspired by the human learning process and propose to enforce the LLMs to deeply understand the problems and pay more attention to the core information relevant to solving the problems. Although the method itself might not introduce too many new technologies, we believe that our findings and approach will promote more related research in the future, which will help the development of this community. Additionally, our method can be easily applied to various LLMs to enhance their reasoning capabilities without introducing any training cost, demonstrating the practical application potential of our approach.

\begin{figure*}[t]
  \centering
  \includegraphics[width=0.85\textwidth]{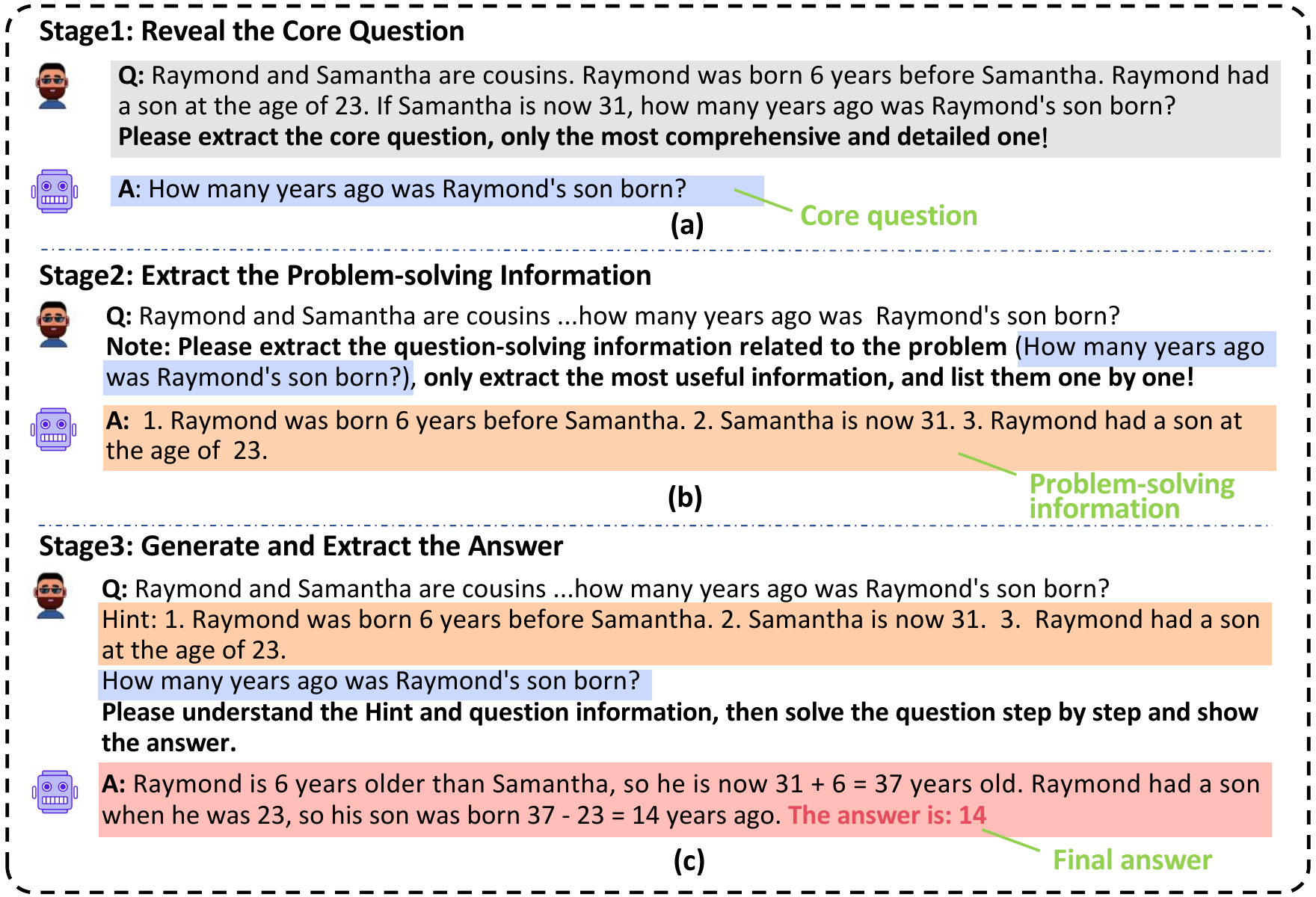}
  \caption{\textbf{Illustration of our DUP prompting strategy}, which contains three-stage processes: \ding{182} revealing the core question from the original input; \ding{183} extracting the problem-solving information based on the core question; \ding{184} generating and extracting the final answer via understanding the core question and problem-solving information.}
  \label{fig_2}
\end{figure*}

\section{DUP Prompting}
\label{sec:method}

\subsection{Overview}
As mentioned in Section~\ref{sec:intro}, semantic misunderstanding is the major error for limiting LLMs' reasoning performance, which has not been well studied in prior works. To this end, we introduce a new zero-shot CoT prompting approach, called DUP prompting, which aims to improve the LLMs' reasoning abilities by enforcing the LLMs to fully understand the problem. Figure~\ref{fig_2} illustrates the process of our DUP method, which contains three-stage processes. Specifically, in stage 1, DUP reveals the \textbf{core question} from a complex and lengthy problem description. In stage 2, DUP further extracts the \textbf{problem-solving information} that is crucial for solving the core question from the same description. In stage 3, given the core question and problem-solving information, DUP incorporates them into the original question to generate the detailed response, and then extracts the \textbf{final answer} from the generated text.

\subsection{Stage 1: Reveal the Core Question}
Understanding the goal of the question is the first step to solving it, even for humans. Unfortunately, LLMs might be confused by lengthy descriptions of complex reasoning questions, leading to inaccurate understanding and poor performance. In response to this problem, we encourage LLMs to explicitly extract the core question from the original input before reasoning. Specifically, we design a core question extraction prompt ``\textbf{\textit{Please extract core question, only extract the most comprehensive and detailed one!}}'', which is appended to the end of question. We then use GPT-3.5-turbo~\cite{ouyang2022training} to extract the core question from the input. As a result, the output of this step will be a shorter and clearer question that will be used to help LLMs focus on the goal of input questions in subsequent steps. 

\subsection{Stage 2: Extract the Problem-solving Information}
In addition to clarifying the goal, it is also important to find the information required to solve the problem. Without fully understanding and utilizing the information provided by the question, reasoning cannot be correctly proceeded. Moreover, it is difficult for LLMs to take full advantage of this information. Therefore, we design a problem-solving information extraction prompt to help solve this problem, \textit{i.e.}, ``\textbf{\textit{Note: Please extract the problem-solving information related to the core question \texttt{[Core Question info]}, only extract the most useful information, list them one by one!}}''. The slot \textbf{\texttt{[Core Question info]}} contains the core question extracted in Stage 1. The output of this step is a list of information, which is useful in reasoning.

\subsection{Stage 3: Generate and Extract the Answer}
Given the extracted core question and problem-solving information extracted in previous stages, we incorporate them into the original input by the template ``\textbf{Hint: \texttt{[Problem-Solving Info]}\textbackslash n\texttt{[Core Question]}\textbackslash n \textit{Please understand the Hint and question information, then solve the problem step by step and show the answer.}}'', where the input slots refer to the corresponding outputs in previous steps. This prompt is beneficial to improve LLMs' understanding of the question by explicitly pointing out the goal and necessary information to solve the question. Lastly, following the prior work~\cite{wang2023planandsolve}, we enforce the LLMs to extract the final numerical answer from the generated long reasoning text. Compared with rule-based matching methods, using LLMs to extract the final answer is more robust and accurate in practice.

\begin{table}[t]
\centering
\scalebox{0.95}{
\begin{tabular}
{lcccr}
\toprule
Dataset & Domain & \# Samples  &Answer Type  \\\midrule
{GSM8K~\cite{gsm8k}}  & Math &1319  &Number  \\
{MultiArith~\cite{mutli_arith}} & Math &600   &Number \\
{AddSub~\cite{addsub}}  & Math &395 &Number  \\
{SVAMP~\cite{svamp}}   & Math &1000  &Number \\
{SingleEq~\cite{singleeq}}  & Math&508  &Number \\
{AQuA~\cite{aqua}} & Math &254   &Option \\
Last Letters~\cite{cot_wei}  & Symbolic &500  & String  \\
Coin Flip~\cite{cot_wei}   & Symbolic &500  & Yes / No \\
{StrategyQA~\cite{strategyqa}}  & Commonsense &2290   &Yes / No   \\
{CSQA~\cite{commonsenseqa}}  & Commonsense &1221  &Option \\

\bottomrule
\end{tabular}
}
\caption{\textbf{Details of all evaluated datasets.} ``Math'', ``Symbolic'' and ``Commonsense'' denote the arithmetic, symbolic and commonsense reasoning, respectively. CSQA refers to the CommonensenseQA benchmark.}
\label{tab:dataset_description}
\end{table}

\section{Experiments}
\label{sec:experiments}
\subsection{Experimental Setup}

\paragraph{\textbf{Tasks and Datasets}}
We conduct extensive experiments on six \textbf{Arithmetic Reasoning} benchmarks, including GSM8K~\cite{gsm8k}, SVAMP~\cite{svamp}, MultiArith~\cite{mutli_arith}, AddSub~\cite{addsub}, AQuA~\cite{aqua} and SingleEq~\cite{singleeq}. Moreover, to investigate the universality of DUP, we also evaluate it on several reasoning tasks in the other domains, \textit{i.e.}, two \textbf{Commonsense Reasoning} benchmarks (CommonsenseQA~\cite{commonsenseqa}, StrategyQA~\cite{strategyqa}) and two \textbf{Symbolic Reasoning} benchmarks (Last Letter~\cite{cot_wei}, Coin Flip~\cite{cot_wei}). The details of all evaluated datasets are shown in Table~\ref{tab:dataset_description}.

\begin{table*}[t]
\centering
\begin{tabular}{lp{11cm}c}
\toprule
No. &Template& Reasoning tasks \\ \midrule
1 &\makecell[l]{
\textbf{Extract core question}: Please extract core question, only the most comprehensive\\ and detailed one!\\
\textbf{Extract problem-solving information }: Note: Please extract the problem-solving\\ information related to the core question [{\sethlcolor{light_blue}\hl{\textit{Core Question}}}], only extract the most\\ useful information, list them one by one!\\
\textbf{Generate the answer}: Hint: [{\sethlcolor{light_orange}\hl{\textit{Problem-solving Info}}}], \textbackslash n[{\sethlcolor{light_blue}\hl{\textit{Core Question}}}].  \textbackslash n Please\\ understand the Hint and question information, then solve the question step by step\\ and show the answer.
 } &{\large \thead{GSM8K~\cite{gsm8k}, AddSub~\cite{addsub}, \\SVAMP~\cite{svamp}, MultiArith~\cite{mutli_arith}, \\SingleEq~\cite{singleeq}, AQuA~\cite{aqua},\\ StrategyQA~\cite{strategyqa}, Coin Flip~\cite{cot_wei},\\
 CommonsenseQA~\cite{commonsenseqa}
 }}\\
\midrule
2 &\makecell[l]{\textbf{Prompt}: Please accurately understand the question useful information and solve the\\ question step by step.} & \small Last Letter~\cite{cot_wei} \\
\bottomrule
\end{tabular}
\caption{\textbf{Reasoning prompt templates of DUP for all reasoning tasks}. Notably, [\textit{Core Question}] indicates the extracted core question, and [\textit{Problem-solving Info}] indicates the extracted problem-solving information to the problem.}
\label{tab:prompt_arithmetic}
\end{table*}

\begin{table*}[t]\centering
\setlength{\tabcolsep}{8pt}
\small
\begin{tabular}{lccccccccc}
\toprule
\multicolumn{1}{c}{}     &       & \multicolumn{6}{c}{\textbf{Arithmetic Reasoning}}          & \multicolumn{2}{c}{\textbf{Score}}\\ 
\cmidrule(lr){3-8} \cmidrule(lr){9-10} 
\multicolumn{1}{c}{\multirow{-2}{*}{\textbf{Model}}} & \multirow{-2}{*}{\textbf{Method}} & \textbf{SVAMP} & \textbf{GSM8K}         & \textbf{AddSub}        & \textbf{MultiArith}    & \textbf{AQuA}          & \textbf{SingleEq}     & \textbf{\textit{\underline{Avg.}}} & \textbf{\textit{$\Delta$ }}  \\ \midrule 
\multicolumn{10}{l}{\textit{Performance of Zero-shot Methods}}  \\ \midrule 
& Zero-shot CoT            & 79.3      & 78.9      & 85.8      & 95.3      & 53.0      & 93.5      & \underline{80.9}          & {\color[HTML]{3CB371}\textbf{ -}}    \\
& Least-to-Most            & 80.9      & 77.5      & 91.3          & 95.5      & 57.4      & 93.5      & \underline{82.6}      & {\color[HTML]{3CB371} +1.7} \\
& Zero-shot PS+   & 80.7      & 79.3      & 86.5      & 92.0      & 55.9      & 93.0      & \underline{81.2}          & {\color[HTML]{3CB371} +0.3 } \\
\multirow{-4}{*}{GPT-3.5-Turbo}             & DUP (Ours)& \textbf{82.5} & \textbf{82.3} & \textbf{92.1} & \textbf{97.8} & \textbf{60.2} & \textbf{94.9} & \textbf{\underline{84.9}} & {\color[HTML]{3CB371}\textbf{+4.0}} \\ \midrule 
\multicolumn{1}{c}{}     & Zero-shot CoT            & 90.4      & 94.6      & 92.4      & 97.8      & 72.8      & 95.0      & \underline{90.6}      & {\color[HTML]{3CB371}\textbf{ -}}    \\
\multicolumn{1}{c}{}     & Least-to-Most            & 90.3      & 92.1      & 92.1      & 97.1      & 71.6      & 95.0      & \underline{89.7}      & {\color[HTML]{FD6864} -0.9} \\
\multicolumn{1}{c}{}     & Zero-shot PS+   & 92.6      & 94.3      & 93.1      & \textbf{98.1} & 75.5      & 95.3      & \underline{91.4}      & {\color[HTML]{3CB371} +0.8} \\
\multicolumn{1}{c}{\multirow{-4}{*}{GPT-4}} & DUP (Ours)                & \textbf{94.2} & \textbf{97.1} & \textbf{95.1} & \textbf{98.1} & \textbf{77.1} & \textbf{96.0} & \textbf{\underline{92.9}} & {\color[HTML]{3CB371} \textbf{+2.3}} \\ \midrule 
\multicolumn{9}{l}{\textit{Performance of Few-shot Methods}}   & \multicolumn{1}{l}{}        \\ \midrule 
& Manual-CoT               & 78.5          & 81.6          & 90.6          & 95.6          & 55.9          & 94.2          & \underline{82.6}          & {\color[HTML]{3CB371} +1.7}    \\
\multirow{-2}{*}{GPT-3.5-Turbo}             & Auto-CoT                 & 82.9          & 80.2          & 89.9          & 99.0          & 54.3          & 94.6          & \underline{83.4}          & {\color[HTML]{3CB371} +2.5}    \\ 

\bottomrule
\end{tabular}
 \caption{\textbf{Results on Arithmetic Reasoning benchmarks}. The best results in the zero-shot setting are in \textbf{bold}. ``\textbf{\textit{$\Delta$ }}'' denotes the average performance {\color[HTML]{3CB371}\textbf{improvement}} or {\color[HTML]{FD6864}\textbf{decline}} of various methods compared to Zero-shot CoT. 
}
 \label{tab:main_results}
\end{table*}

\paragraph{\textbf{Compared Methods}} 
Since our DUP is a zero-shot prompting method, we mainly compare it with other zero-shot methods. For references, two typical few-shot prompting methods are also used as the baselines.
\begin{itemize}
    \item Zero-shot CoT~\cite{kojima2022large} simply adds a prompt ``Let’s think step by step'' before each answer.
    \item Least-to-Most~\cite{zhou2023leasttomost} aims to break down a complex problem into a series of simpler sub-problems and then solve them in sequence.
    \item Plan-and-Solve~\cite{wang2023planandsolve} devises a plan to divide the entire task into smaller sub-tasks, and then carries out the sub-tasks according to the plan. Specifically, we adopt the more sophisticated Plan-and-Solve (PS+) prompting method in this work.
    \item Manual-CoT~\cite{cot_wei} is the first CoT method that proposes to use a few CoT demonstrations as exemplars in prompting.
    \item Auto-CoT~\cite{zhang2022automatic} improves the vanilla CoT via sampling questions with diversity and generating reasoning chains to construct demonstrations.
\end{itemize}

\paragraph{\textbf{Implementation Details}}
We use the public GPT-3.5-Turbo (0613)~\cite{ouyang2022training} and GPT-4 (0613)~\cite{openai2023gpt4} as the test LLMs. 
For the implementation of our DUP, we show the reasoning prompt templates for all reasoning tasks in Table~\ref{tab:prompt_arithmetic}. Notably, since the Last Letter~\cite{cot_wei} is a simple symbolic reasoning task that does not rely on the problem-solving information, we skip the stage-2 process of DUP and use a simplified prompt. In this work, all models are employed via OpenAI's API, and we adopt the greedy decoding strategy with the temperature setting of 0 across all experiments. For the few-shot prompting baselines, we keep the recommended number of demonstration examples specified in their original papers.

\subsection{Main Results}

\paragraph{\textbf{Arithmetic Reasoning}} Table~\ref{tab:main_results} presents the main results of Arithmetic Reasoning benchmarks. As seen, compared to the vanilla zero-shot CoT, our DUP method brings consistent and significant performance gains across all reasoning benchmarks. Specifically, in GPT-3.5-turbo settings, DUP improves the accuracy by an average of 4\% over Zero-shot CoT. When using GPT-4, our DUP even achieves new state-of-the-art results on \textbf{GSM8K (97.1\%)} and \textbf{SVAMP (94.2\%)}, indicating its superiority.

Moreover, we also report the results of few-shot counterparts. Due to the high cost of GPT-4 API, we use the more affordable GPT-3.5-turbo as the responder for few-shot methods. Generally, the performance of zero-shot methods tends to be lower than that of few-shot methods. However, with the help of DUP, GPT-3.5 can even achieve remarkable zero-shot performance that is higher than few-shot methods. These results prove the effectiveness of DUP.

\begin{table}[t]
    \centering
    \setlength{\tabcolsep}{5pt}
    \begin{tabular}{lcccc}
    \toprule
     \textbf{Method}& \textbf{CSQA} & \textbf{StrategyQA}& \textbf{\textit{\underline{Avg.}}} & $\Delta$ \\    \midrule
    Zero-shot CoT & 72.3 & 66.1 & \underline{69.2} &-  \\
    Least-to-Most & 71.9 & 61.5 & \underline{66.7} & {\color[HTML]{FD6864} -2.5} \\
    Zero-shot PS+ & 68.8 & 62.8 & \underline{65.8} & {\color[HTML]{FD6864} -3.4} \\
    DUP (Ours) &\textbf{74.5} &  \textbf{68.5} & \textbf{\underline{71.5}} & {\color[HTML]{3CB371} \textbf{+2.3}} \\
    \midrule
    Few-shot Manual-CoT & 76.5 & 64.8 & \underline{70.8} & {\color[HTML]{3CB371} +1.6} \\
    Few-shot Auto-CoT & 74.2 & 62.5 & \underline{68.3} & {\color[HTML]{FD6864} -0.9} \\
    \bottomrule
    \end{tabular}
    \caption{\textbf{Results of GPT-3.5-turbo on Commonsense Reasoning}.}
    \label{tab:cs}
\end{table}

\begin{table}[t]
    \centering
    \setlength{\tabcolsep}{3pt}
    \begin{tabular}{lccccc}
    \toprule
     \textbf{Method}& \textbf{Last Letter} & \textbf{Coin Flip}& \textbf{\textit{\underline{Avg.}}} &$\Delta$ \\ \midrule
    Zero-shot CoT & 60.8 & 94.4 & \underline{77.6} &-\\
    Least-to-Most & \textbf{83.2} & 82.8 & \underline{83.0} &{\color[HTML]{3CB371} +2.4} \\
    Zero-shot PS+ & 60.6 & 95.4 & \underline{78.0} &{\color[HTML]{3CB371} +0.4}\\
    DUP (Ours) & 81.2 &  \textbf{97.6} & \textbf{\underline{89.4}} &{\color[HTML]{3CB371} \textbf{+11.8}} \\
    \midrule
    Few-shot Manual-CoT & 74.4 & 98.2 & \underline{86.3} &{\color[HTML]{3CB371} +8.7}\\
    Few-shot Auto-CoT & 81.2 & 98.6 & \underline{89.9} &{\color[HTML]{3CB371} +12.3} \\
    \bottomrule
    \end{tabular}
    \caption{\textbf{Results of GPT-3.5-turbo on Symbolic Reasoning}.}
    \label{tab:sys_reasoning}
\end{table}

\paragraph{\textbf{Commonsense and Symbolic Reasoning}}
Table~\ref{tab:cs} shows the performance on Commonsense Reasoning datasets. Considering the experimental cost, we only used GPT-3.5-turbo as the test LLM. Compared to zero-shot methods, DUP consistently outperforms all counterparts. In comparison with few-shot methods, DUP also achieves comparable or even better performance. Table~\ref{tab:sys_reasoning} lists the results on Symbolic Reasoning datasets. On Last Letters, zero-shot DUP (81.2\%) is marginally worse than Zero-shot Least-to-Most (83.2\%), on par with few-shot Auto-CoT (81.2\%), but significantly exceeds other zero-shot methods and few-shot Manual-CoT (74.4\%). On Coin Flip, zero-shot DUP (97.6\%) is slightly worse than few-shot Manual-CoT (98.2\%) and few-shot Auto-CoT (98.6\%), but significantly outperforms other zero-shot baseline methods. In general, we can basically conclude that DUP outperforms other zero-shot methods, and has great potential to beat the few-shot methods.

\begin{table}[t]
\centering
\setlength{\tabcolsep}{7pt}
\begin{tabular}{cccccc}
\toprule
\textbf{Stage 1} & \textbf{Stage 2} & \textbf{Stage 3} & \textbf{GSM8K} & \textbf{AQuA} & \textbf{\textit{\underline{Avg.}}} \\
\midrule
\usym{2717}      & \usym{2717}    & \usym{2717}    & 76.5          & 51.2          & \underline{63.8}          \\
\usym{2713}      & \usym{2717}    & \usym{2717}    & 78.9           & 53.1          & \underline{66.0}            \\
\usym{2717}    & \usym{2713}      & \usym{2717}    & 80.6           & 55.1          & \underline{67.8}            \\
\usym{2717}    & \usym{2717}    & \usym{2713}      & 80.3           & 54.7          & \underline{67.5}            \\
\usym{2713}      & \usym{2713}      & \usym{2717}    & 79.9           & 57.0          & \underline{68.4}            \\
\usym{2713}      & \usym{2717}    & \usym{2713}      & 80.8           & 56.2          & \underline{68.5}            \\
\usym{2717}    & \usym{2713}      & \usym{2713}      & 81.7           & 58.2          & \underline{69.9}            \\
\usym{2713}      & \usym{2713}      & \usym{2713}      & \textbf{82.3}  & \textbf{60.2} & \textbf\underline{{71.2}}            \\
\bottomrule
\end{tabular}
\caption{\textbf{Ablation study for different variations of DUP}. We report the results of GPT-3.5-turbo on GSM8K and AQuA. Notably, Stage 1 involves extracting core questions, Stage 2 focuses on extracting problem-solving information, and Stage 3 entails solving the problem step by step. }
\vspace*{-10pt} 
\label{tab:ablation}
\end{table}

\subsection{Ablation Study}
In this part, we conduct a series of ablation experiments to investigate the impact of each stage in our DUP, where Stage 1 of DUP involves extracting core questions, Stage 2 focuses on extracting problem-solving information, and Stage 3 entails solving the problem step by step. In Table~\ref{tab:ablation}, we report the results of various combinations of the three stages in our DUP. Specifically, we conduct the ablation experiments on GPT-3.5-turbo and present the results of GSM8K and AQuA benchmarks. As seen, removing each stage results in performance degradation, and the combination of all stages achieves the best performance on both benchmarks. These results demonstrate the importance of each stage in our DUP.

\begin{figure}[t]
  \centering
    \includegraphics[ width=\linewidth ]{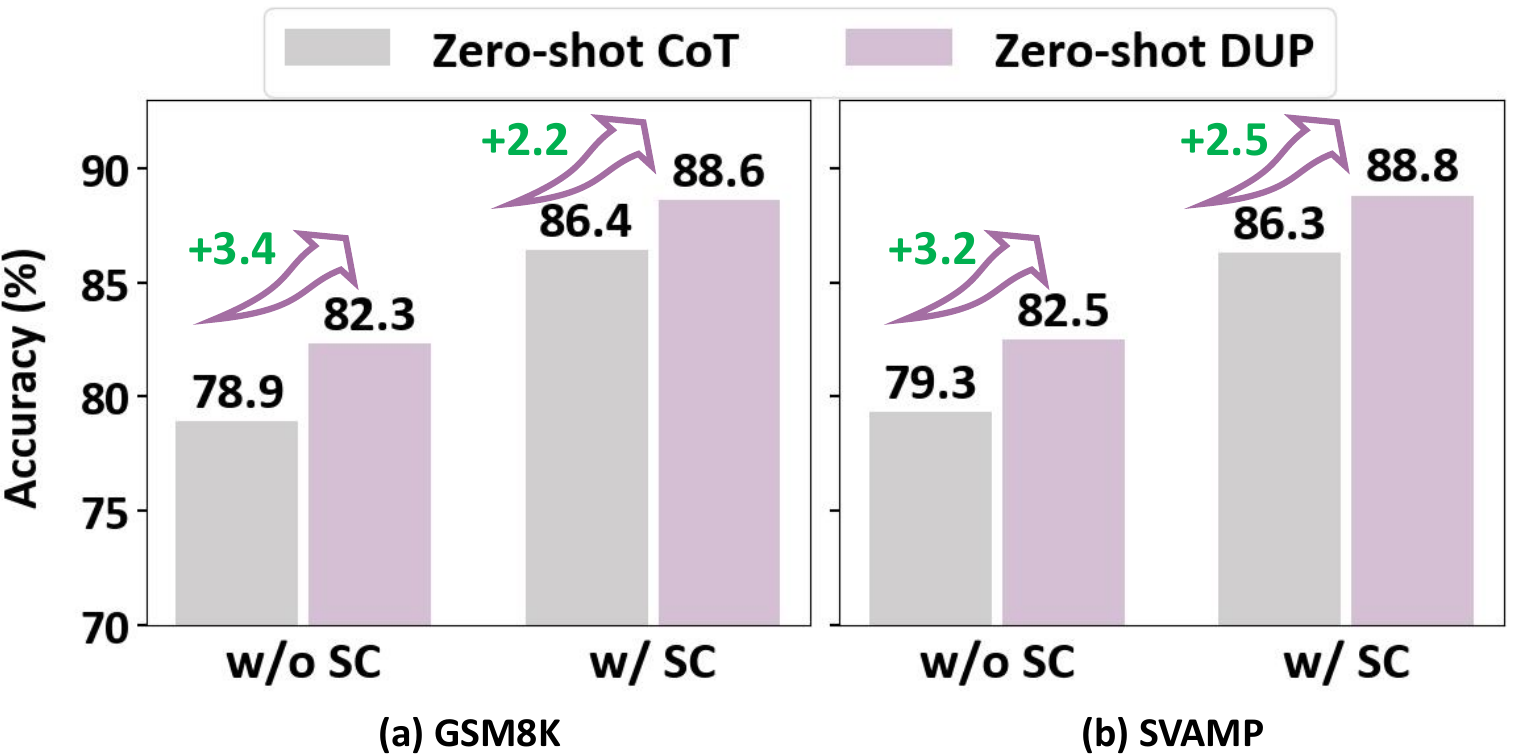}
  \caption{
   \textbf{Results of DUP Prompting with and without self-consistency(SC) }using GPT-3.5-turbo LLM on GSM8K and SVAMP.
  }
  \label{fig:sc}
\end{figure}

\subsection{Discussion}
\paragraph{\textbf{Compatibility with Self-consistency}}
We employ an innovative decoding strategy with self-consistency (SC)~\cite{wang2022self_consistency} as a substitute for the conventional greedy decoding approach, which initially samples \textit{N} reasoning paths rather than only opting for the greedy approach. Subsequently, choosing the most consistent answer as the answer. Existing works~\cite{wang2023planandsolve,xu2023re} indicate that adopted SC notably enhances the performance of chain-of-thought prompting. Here, to verify whether using SC can further enhance the performance of DUP, we conduct experiments on GSM8K and SVAMP using GPT-3.5-Turbo, setting the temperature to 0.7 and \textit{N} to 10. The results are illustrated in Figure~\ref{fig:sc}, where the SC strategy brings remarkable performance improvements. Notably, DUP with SC (88.6\% and 88.8\%) consistently outperforms Zero-shot CoT with SC (86.4\% and 86.3\%), continuing to prove its superiority.

\begin{figure*}[t]
  \centering
  \includegraphics[width=0.9\linewidth ]{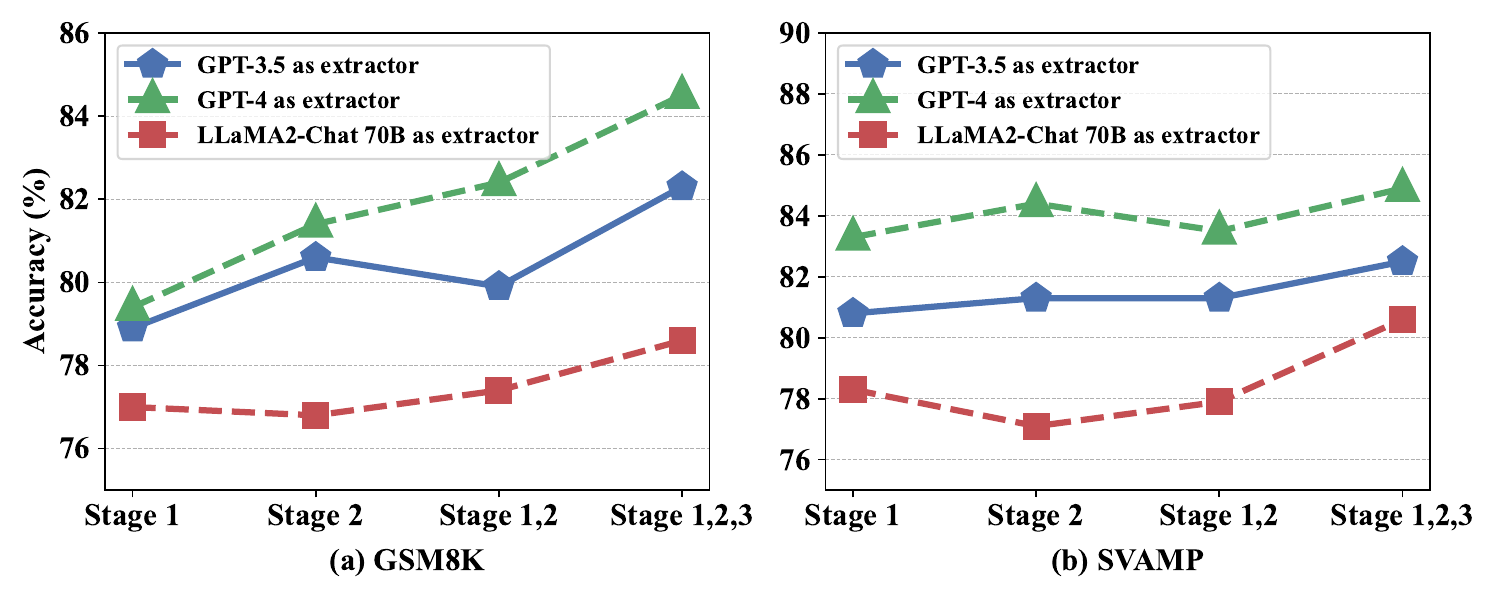}
  \caption{
    \textbf{Analysis of different information extractors used in our DUP}. We use the GPT-4, GPT-3.5-turbo, and Llama-2-Chat 70b to extract core question (Stage1) and problem-solving information (Stage2) extractor, and leverage the extracted contents to guide the responses of GPT-3.5-turbo (Stage3). We see that more accurate core questions and problem-solving information lead to better performance.
  }
  
  \label{fig:extractor}
\end{figure*}

\paragraph{\textbf{More Accurate Core Questions and Problem-solving Information Lead to Better Performance}}
As stated in Section~\ref{sec:intro}, the core of our DUP is to guide LLMs to deeply understand the problems, \textit{i.e.}, extracting the core question and key problem-solving information. To verify it, we conduct contrastive experiments on AQuA, GSM8K, and SVAMP datasets. Specifically, using the GPT-3.5-Turbo as the final responder, we leverage different LLMs (\textit{i.e.}, LLaMA2-Chat-70B, GPT-3.5, GPT-4) to extract the core question in Stage 1 and the key problem-solving information in Stage 2, respectively. The contrastive results are illustrated in Figure~\ref{fig:extractor}. As seen, when using the GPT-4 as the extractor, GPT-3.5 responder can achieve better performance than that using GPT-3.5 as the extractor. Conversely, using the LLaMA2-Chat-70B as the extractor leads to worse results. These results demonstrate that better core questions and key problem-solving information can result in better reasoning performance, confirming our statement.

\begin{figure}[t]
  \centering
  \includegraphics[width=1\linewidth]{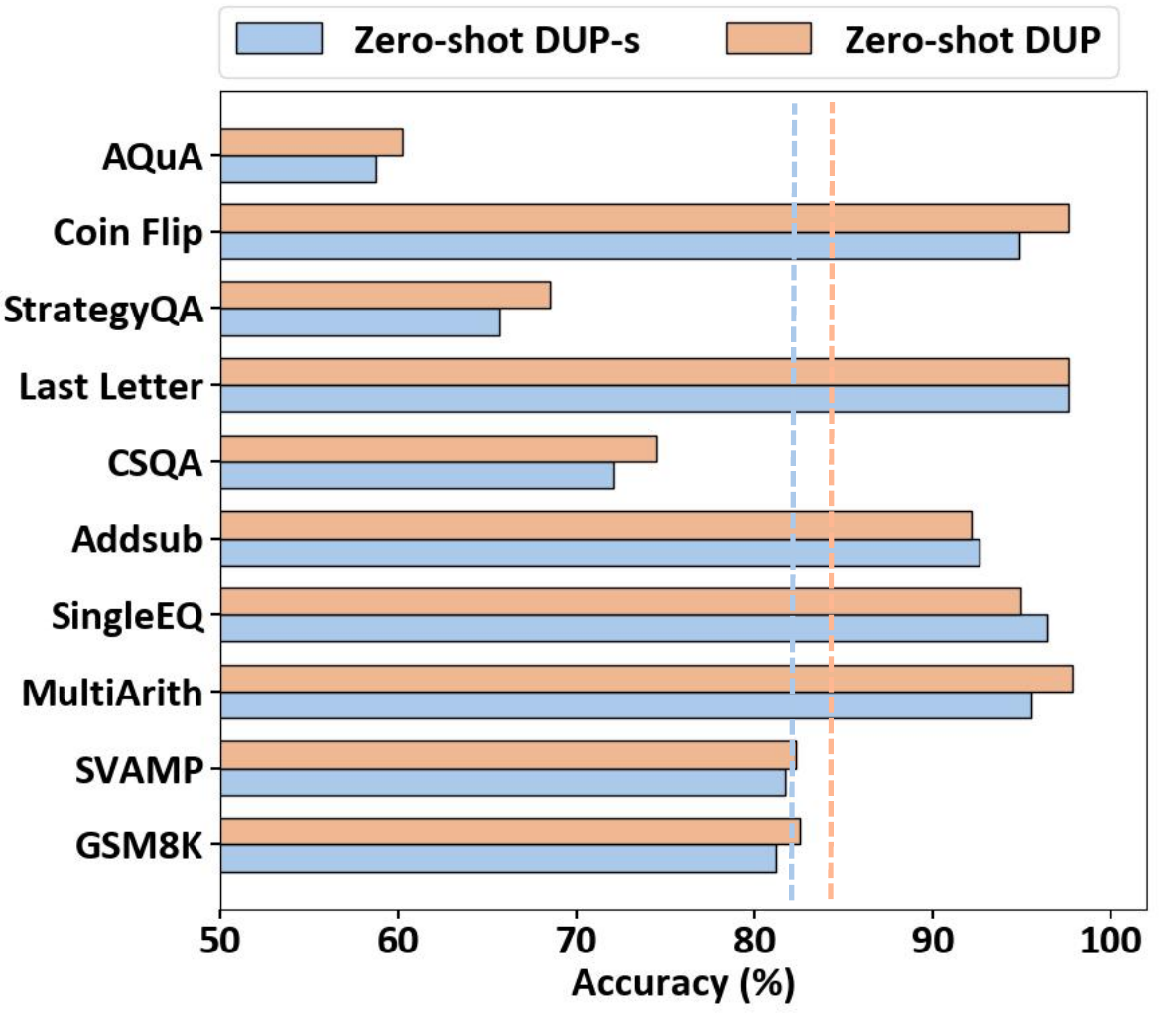}
  \caption{
    \textbf{Performance of DUP and DUP-s across various reasoning tasks on GPT-3.5-Turbo}, where DUP-s merges the three-stage prompts into one prompt. {\color[HTML]{EFB792}\textbf{Orange}} and {\color[HTML]{ABC9EA}\textbf{Blue}} dashlines represent the average accuracy of DUP and DUP-s, respectively. We see that our simplified DUP-s also achieves remarkable performance with less inference budget.
  }  
  \label{fig:reduce}
\end{figure}

\paragraph{\textbf{Reduce inference cost without much performance degradation}}
Some readers may be concerned that the three-stage processes in DUP will cause too much inference cost. Hence, we further propose the simplified DUP method, namely DUP-s, which merges the three-stage prompts into one prompt. We conduct contrastive experiments on all 10 reasoning benchmarks, and illustrate the results in Figure~\ref{fig:reduce}. As seen, DUP-s achieves comparable performance to DUP, and even achieves better performance on two tasks of Addsub and SingleEQ. Therefore, in the case of a limited inference budget, using our simplified DUP-s method is also a good choice.

\begin{figure*}[t]
  \centering
   \includegraphics[width=0.95\linewidth ]{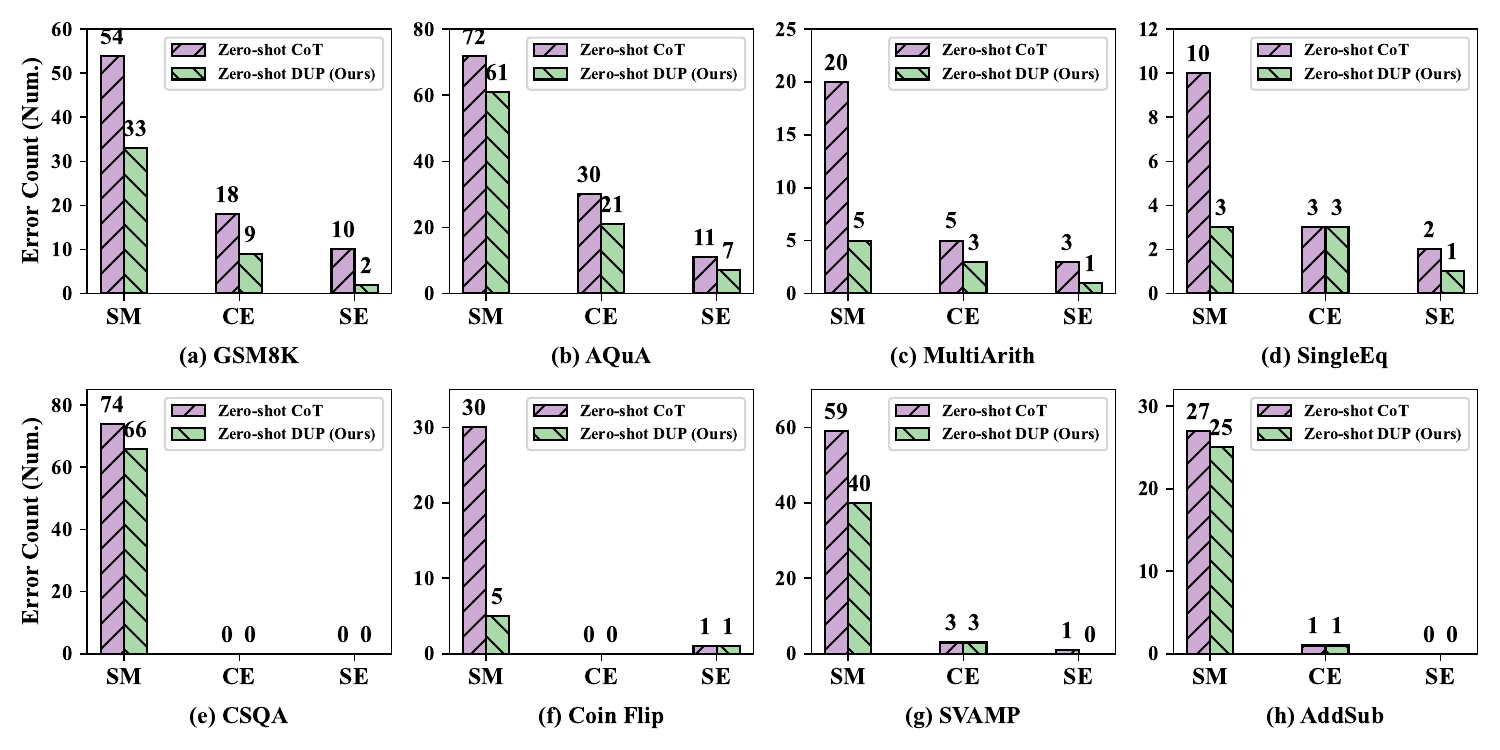}
     \caption{\textbf{Quantitative error analyses of different prompting methods.} Notably, ``SM'', ``CE'' and ``SE'' denote the ``Semantic Misunderstanding'', ``Calculation Error'' and ``Step-missing Error''.
     We randomly select 300 examples for each reasoning dataset (except AQuA which only contains 254 examples), and use GPT-3.5-Turbo LLM to generate responses and count failed answers.
     We can see that our method reduces the frequency of various error types compared with Zero-shot CoT.}
  \label{fig:error_analysis}
\end{figure*}

\begin{table}[t]
    \centering
    \setlength{\tabcolsep}{11pt}
    \scalebox{0.95}{
    \begin{tabular}{lcccc}
    \toprule
    \textbf{Method} & \textbf{GSM8K} & \textbf{AddSub} & \textbf{\underline{Avg.}} &$\Delta$ \\  \midrule
    \multicolumn{5}{l}{\textit{LLaMA-2-Chat-13B}} \\ \hdashline
    Zero-shot CoT    & 35.1   & 70.6    & \underline{52.8}   &-       \\
    DUP (Ours)   & \textbf{35.9} & \textbf{79.7}   & \textbf{\underline{57.8}} & {\color[HTML]{3CB371}\textbf{+5.0}}    \\  \midrule
        \multicolumn{5}{l}{\textit{LLaMA-2-Chat-70B}} \\ \hdashline
    Zero-shot CoT    & 53.9   & 75.6    & \underline{64.7}   &-       \\
    DUP (Ours)   & \textbf{56.4} & \textbf{87.8}   & \textbf{\underline{72.1}} & {\color[HTML]{3CB371}\textbf{+7.4}}    \\  \midrule
        \multicolumn{5}{l}{\textit{CodeLLaMA-Instruct-13B}} \\ \hdashline
    Zero-shot CoT    & 24.2   & 73.1    & \underline{48.6}   &-       \\
    DUP (Ours)   & \textbf{28.1} & \textbf{74.6}   & \textbf{\underline{51.3}} & {\color[HTML]{3CB371}\textbf{+2.7}}    \\  \midrule
        \multicolumn{5}{l}{\textit{CodeLLaMA-Instruct-34B}} \\ \hdashline
    Zero-shot CoT    & 39.1   & 81.2    & \underline{60.1}   &-       \\
    DUP (Ours)   & \textbf{43.5} & \textbf{86.0}   & \textbf{\underline{64.7}} & {\color[HTML]{3CB371}\textbf{+4.1}}    \\  \bottomrule
    \end{tabular}
}
\caption{\textbf{Results of various Open-source LLMs using DUP on GSM8K and Addsub.} We see that DUP still outperforms Zero-shot CoT by a clear margin.}
\label{tab:samll_result}
\end{table}

\paragraph{\textbf{Whether DUP also works well on Open-source LLMs}}
In the above experiments, we mainly evaluate our DUP in the close-source GPT LLMs. To verify whether our DUP also works well on other open-source LLMs, we evaluate our method on 4 widely-used LLMs, \textit{i.e.}, LLaMA-2-Chat 13B and 70B models~\cite{touvron2023llama}, CodeLLaMA-Instruct 13B and 34B models~\cite{roziere2023code}. As seen in Table~\ref{tab:samll_result}, in the cases of open-source LLMs, our DUP can still outperform the baseline zero-shot CoT by a large margin on GSM8K and AddSub benchmarks. This also proves the universality of our proposed DUP method.

\begin{table}[t]
\centering
\small
\scalebox{0.85}{
\begin{tabular}{p{9.5cm}}
\toprule
Template \\ \midrule
\makecell[l]{
\textbf{Question}: [{\sethlcolor{light_blue}\hl{\textit{Input Question}}}].\\
\textbf{Wrong Response}: [{\sethlcolor{light_red}\hl{\textit{Wrong Answer}}}].\\
\textbf{Correct Response}: [{\sethlcolor{light_red}\hl{\textit{Correct Answer}}}].\\
Please judge which type of error it belongs to based on the above information:\\
\quad 1. Semantic Misunderstanding: semantic misunderstanding or lack of\\ commonsense concepts.\\
\quad 2. Calculation error: errors occurred while performing a basic
operation.\\
\quad 3. Step-missing errors: missing step and hallucination.\\
Finally, please explain why this error falls into the category you
select.
 } 
\\ \bottomrule
\end{tabular}
}
\caption{\textbf{Prompts for error analysis.} The slot [\textit{Input Question}] denotes the original problem. The slots [\textit{Wrong Question}] and [\textit{Correct Question}] denote the incorrect text generated by the LLMs and the original label.}
\label{tab:prompt_error}
\end{table}

\begin{figure*}[ht]
  \centering
  	\includegraphics[width=0.95\linewidth]{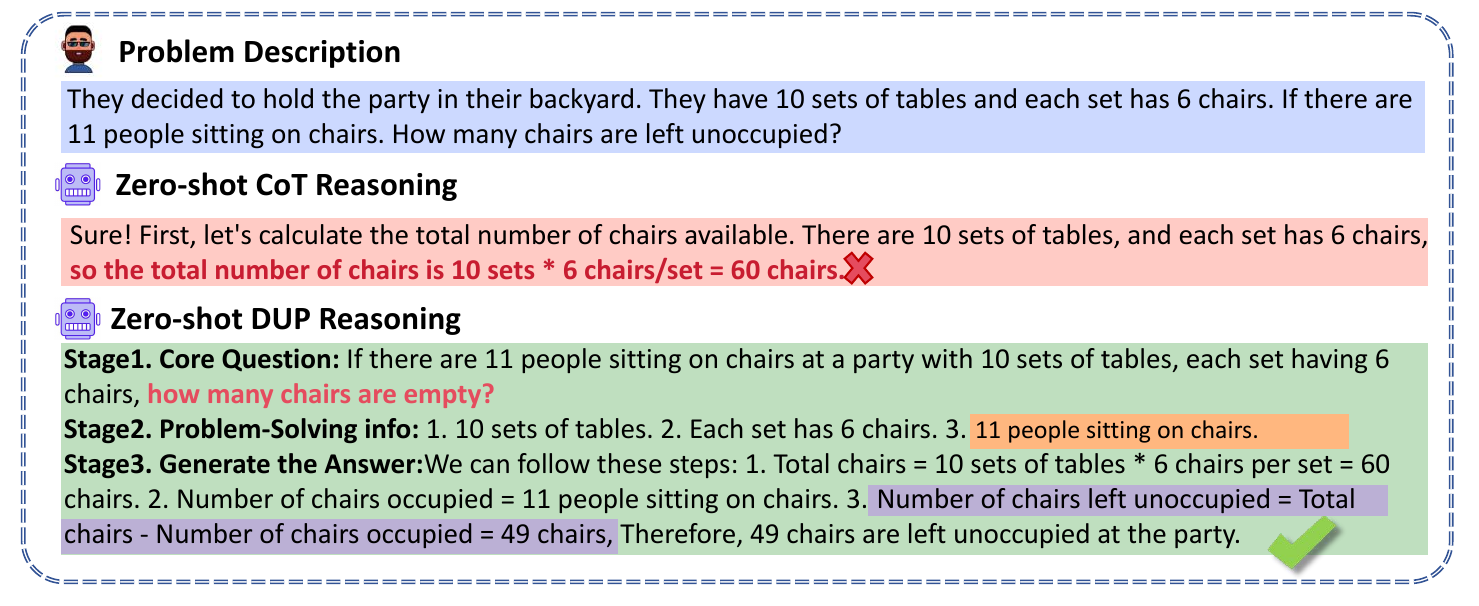} 
  \caption{
     \textbf{Case study on SVAMP.} Zero-shot CoT fails to generate the correct answer, but our DUP method can extract more related problem-solving information and make the correct prediction via deeply understanding the problems.
  }
  \label{fig:case_study}
\end{figure*}

\subsection{Error Analysis}
Here, to verify whether DUP indeed reduces the semantic misunderstanding, we randomly select 300 samples for each reasoning dataset, and perform error analysis for the questions with incorrect answers by prompting the GPT-3.5 LLM. The detailed prompt used to categorize the failure examples is shown in Table~\ref{tab:prompt_error} and the quantitative results are illustrated in Figure~\ref{fig:error_analysis}. As seen, compared with the baseline zero-shot CoT, our DUP reduces semantic misunderstanding effectively. Additionally, we can also find that DUP reduces the calculation and step-missing error as well. One possible reason is that learning more problem-solving information can lead to more accurate reasoning steps. 

Moreover, some readers may concern whether the above LLM-based error analyses are reliable and trustworthy. Regarding this concern, we would like to state that, although LLMs may not have the ability to directly solve the difficult reasoning problems, LLMs can be instructed to reflect on their own CoT, which allows them to identify errors and explain the cause of these errors~\cite{madaan2024self}. Such a self-reflect ability of LLMs has been explored by many prior studies~\cite{Yao2023TreeOT,pan2024automatically}. To further verify the credibility of this LLM-based analysis method, we manually classify the error types for several benchmarks. The contrastive results are listed in Table~\ref{tab:manual_analsis}. As seen, there is a significant correlation between the manual analysis results and the LLM-based results, validating the reliability of our analyses.

\begin{table*}[t]
\centering
\setlength{\tabcolsep}{12pt}
\begin{tabular}{lcccccc}
\toprule
\multicolumn{1}{c}{\multirow{2}{*}{\textbf{Error type}}} & \multicolumn{2}{c}{\textbf{GSM8K}} & \multicolumn{2}{c}{\textbf{MultiArith}} & \multicolumn{2}{c}{\textbf{Coin Filp}} \\ \cmidrule(lr){2-3} \cmidrule(lr){4-5} \cmidrule(lr){6-7}
\multicolumn{1}{c}{} & LLM-based & Manual & LLM-based & Manual & LLM-based & Manual \\ \midrule
Semantic Misunderstanding &54 / 33 &64 / 39  &20 / 5  &24 / 7  & 30 / 5 &31 / 5  \\
Calculation   Error &18 / 9  &6 / 2	&5 / 3	&0 / 0	&0 / 0	&0 / 0  \\
Step-missing   Error & 10 / 2 &12 / 3	&3 / 1	&4 / 2	&1 / 0	&0 / 0 \\
\bottomrule
\end{tabular}
\caption{\textbf{Comparison between LLM-based and manual error analysis methods.} Notably, we report the analysis results on GPT-3.5’s responses (as mentioned in Figure~\ref{fig:error_analysis}). For each result, we list the number of errors prompting with zero-shot CoT and our DUP, respectively.}
\label{tab:manual_analsis}
\end{table*}

To have a close look, we present a case study on SVAMP, as shown in Figure~\ref{fig:case_study}. It can be seen that the zero-shot CoT fails to generate the correct answer, but with the help of our DUP, the LLMs can better understand the problems and generate an accurate answer. 

\section{Conclusion}
\label{sec:conclusion}
In this work, we reveal that deeply understanding the whole problem is crucial for tackling complex reasoning tasks. Consequently, we introduce the DUP method to improve the LLMs' reasoning abilities by encouraging them to deeply understand the problem. A series of experiments on arithmetic, commonsense, and symbolic reasoning tasks prove that DUP brings consistent and significant performance gains across all benchmarks and LLMs. Additionally, DUP outperforms the other zero-shot counterparts by a large margin, and achieves new SOTA results in two popular benchmarks, \textit{i.e.}, GSM8K and SVAMP. More in-depth discussions and systematic analyses further reveal when and where our DUP works well. Moreover, considering that fully understanding the whole problem may also be beneficial to non-reasoning tasks, we will attempt to expand our method to more fields in future work.

\begin{acknowledgement}
This work was supported in part by the National Key Research and Development Program of China under Grant 2023YFC2705700, in part by the National Natural Science Foundation of China under Grants 623B2076, U23B2048, 62076186 and 62225113, and in part by the Innovative Research Group Project of Hubei Province under Grant 2024AFA017. The numerical calculations in this paper have been done on the supercomputing system in the Supercomputing Center of Wuhan University.
\end{acknowledgement}

\begin{competinginterest}
The authors declare that they have no competing interests or financial conflicts to disclose.
\end{competinginterest}

\bibliographystyle{fcs}
\bibliography{ref}

\end{document}